\documentclass[10pt,letterpaper,twocolumn]{article}

\usepackage[T1]{fontenc}
\usepackage{lmodern}
\usepackage[margin=0.72in]{geometry}
\usepackage{microtype}
\usepackage{graphicx}
\usepackage{booktabs}
\usepackage{tabularx}
\usepackage{amsmath}
\usepackage{amssymb}
\usepackage{caption}
\usepackage{placeins}
\usepackage[round,authoryear]{natbib}
\usepackage[hyphens]{url}
\usepackage[hidelinks]{hyperref}

\hypersetup{
  pdftitle={One Geometry, Different Outcomes: Readout-Dependent Effects of the Modality Gap in Vision-Language Models},
  pdfauthor={Aditya Sharma and Divya Saxena}
}

\makeatletter
\renewcommand\section{\@startsection{section}{1}{\z@}%
  {-3.5ex \@plus -1ex \@minus -.2ex}%
  {2.3ex \@plus .2ex}%
  {\normalfont\Large\bfseries\raggedright\hyphenpenalty=10000\exhyphenpenalty=10000}}
\renewcommand\subsection{\@startsection{subsection}{2}{\z@}%
  {-3.25ex\@plus -1ex \@minus -.2ex}%
  {1.5ex \@plus .2ex}%
  {\normalfont\large\bfseries\raggedright\hyphenpenalty=10000\exhyphenpenalty=10000}}
\AtBeginDocument{%
  \setlength{\@fptop}{0pt}%
  \setlength{\@fpsep}{12pt}%
  \setlength{\@fpbot}{0pt plus 1fil}%
  \setcounter{topnumber}{4}%
  \setcounter{dbltopnumber}{2}%
  \setcounter{totalnumber}{6}%
}
\makeatother

\title{One Geometry, Different Outcomes: Readout-Dependent Effects of the Modality Gap in Vision-Language Models}
\author{Aditya Sharma \quad Divya Saxena\\
\small Indian Institute of Technology Jodhpur\\
\small \texttt{p25ai0201@iitj.ac.in} \quad \texttt{divyasaxena@iitj.ac.in}}
\date{}

\begin{document}
\raggedbottom
\maketitle

\begin{abstract}
Contrastive vision--language models learn shared embedding spaces by aligning matched image--text pairs, yet their representations remain separated by a modality gap. Prior work reports divergent effects of modifying this gap: reducing it can improve zero-shot classification and cross-modal alignment, whereas removing gap-related structure can degrade image--text retrieval. In this paper, we provide a unified geometric explanation for these task-dependent effects.
Across CLIP and SigLIP encoders, we find that a single dominant direction captures 94.4--99.9\% of the squared norm of the image--text mean separation, revealing that the mean-separation component is approximately rank-one. A decomposition of the similarity score then identifies three task-specific roles.
In zero-shot classification, query-side fixed gap-offset subtraction is exactly equivalent to an additive class bias. In standard cross-modal retrieval, projecting out the gap direction and renormalising residuals discards candidate-specific norm information, inducing a multiplicative ranking distortion; a geometry-derived exponent tracks the grid-search optimum (Spearman $\rho=0.93$) and restores performance in some settings, although the gains transfer unevenly. In mixed-modal retrieval, the gap direction sorts candidates by modality; its removal can improve cross-modal ranking, unlike random or non-gap controls. Residual semantic structure after removal defines the limits of the rank-one account. Together, these results explain why gap modification can improve, degrade, or restore performance across downstream settings.
By clarifying when and why gap modification changes model behavior, this account provides a principled basis for selecting gap interventions in similarity-based vision--language systems across evaluated downstream tasks.
\end{abstract}

\section{Introduction}
\label{sec:introduction}

Contrastive vision-language models such as CLIP and SigLIP align matched image--text pairs in a shared representation space~\citep{radford2021learning,zhai2023sigmoid} that supports zero-shot classification, cross-modal retrieval, and multimodal search. Yet their image and text embeddings remain systematically separated, forming a persistent modality gap~\citep{liang2022mind}. Prior studies report conflicting outcomes: reducing this separation can improve selected cross-modal tasks, whereas stronger alignment or removal of modality-specific structure can degrade classification or standard image--text retrieval. This disagreement arises partly because the gap is often treated as though it has a task-independent effect. Figure~\ref{fig:empirical-motivation} summarizes how the same dominant mean-separation direction produces different effects under different operations and downstream readouts.

\begin{center}
  \centering
  \includegraphics[width=\columnwidth]{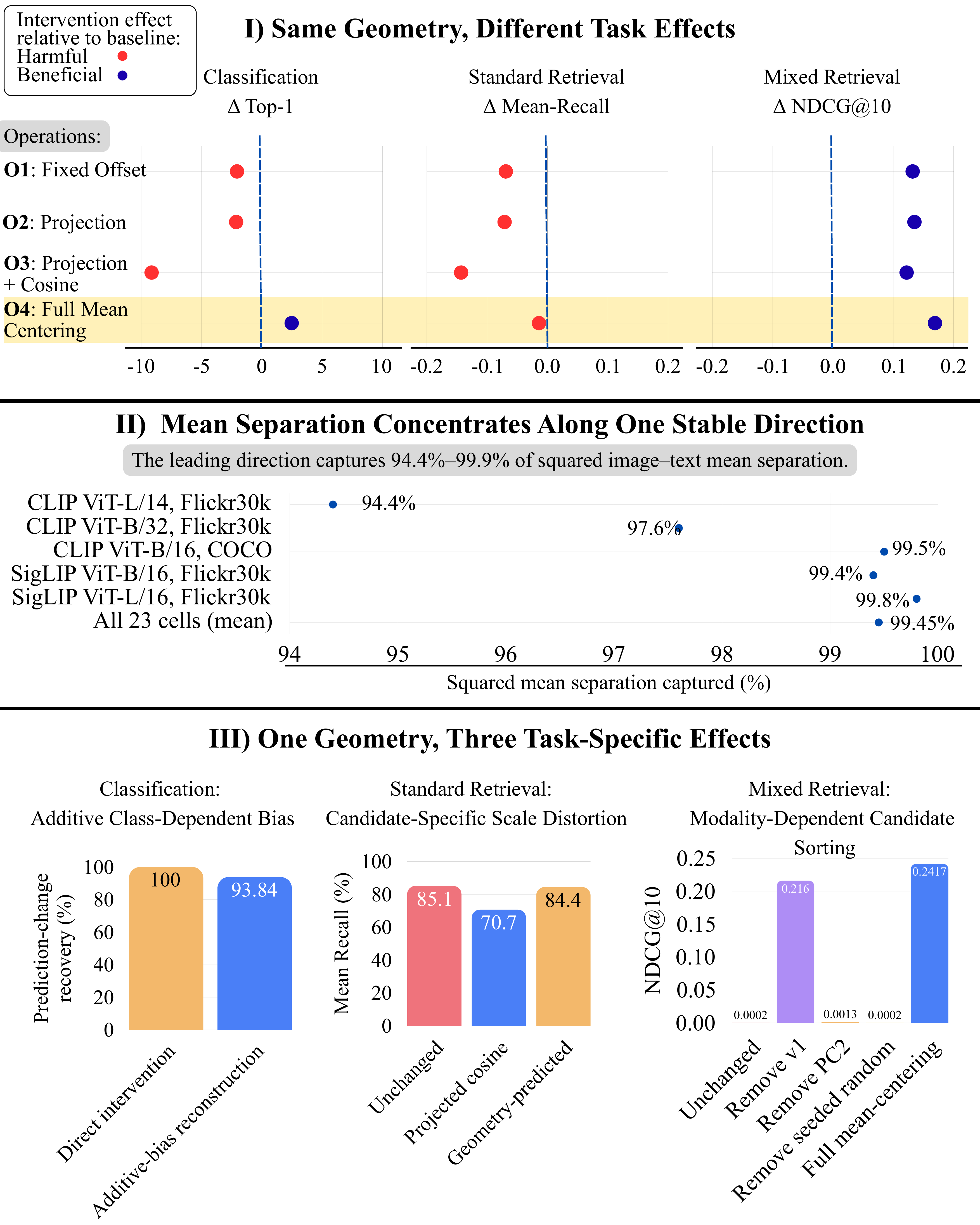}
  \captionof{figure}{\textbf{One dominant mean-separation direction, readout-dependent outcomes.} Panel I compares O1--O4 on task-specific effect scales over 15 classification, 16 standard-retrieval, and 16 matched mixed-retrieval configurations with five paired seeds. Panel II shows representative fractions of squared centroid separation aligned with $v_1$. Panel III summarizes the classification, standard-retrieval, and mixed-retrieval mechanism checks.}
  \label{fig:empirical-motivation}
\end{center}

Two distinctions resolve the apparent conflict. First, interventions are non-equivalent: fixed-offset subtraction, orthogonal projection, projection followed by L2 normalization, full modality-wise mean-centering, and score calibration remove different information. Second, downstream readouts hold different quantities fixed. Classification compares one image with varying text prototypes; standard cross-modal retrieval ranks opposite-modality candidates; mixed-modal retrieval lets images and text compete in one index. An explanation must therefore specify both the geometric operation and the readout.

We show that image--text mean separation concentrates along a stable leading joint-cloud direction across CLIP and SigLIP encoders. Under fixed-query classification, this coordinate induces an additive class-dependent bias. After projection and normalization in standard retrieval, discarded candidate-specific residual-length information produces a multiplicative scale distortion. In a mixed index, it sorts candidates by modality. The same component can therefore help or harm depending on how the downstream system reads the space.

Across 23 encoder--dataset cells, the leading direction captures 94.4--99.9\% of the squared image--text mean separation and remains stable across splits, seeds, datasets, and encoders. Analytical reconstruction, matched controls, and frozen estimate/evaluation protocols support the readout-dependent mechanisms. Held-out transfer preserves qualitative effects more reliably than quantitative magnitudes, while residual multidimensional structure defines the boundary of the account.

Our contributions are as follows:
\begin{itemize}
  \item \textbf{Dominant mean-separation geometry.} We show that 94.4--99.9\% of squared image--text centroid separation aligns with one stable joint-cloud direction across 23 CLIP and SigLIP encoder--dataset cells, while matched non-gap axes do not reproduce its modality separation.
  \item \textbf{Readout-dependent mechanism.} We derive a common score decomposition linking the same coordinate to additive class bias, candidate-specific scale distortion, and modality-dependent sorting under three downstream readouts.
  \item \textbf{Mechanism validation and scope.} Analytical reconstruction, matched direction controls, configuration-specific retrieval calibration, and a frozen held-out checkpoint establish the explanatory range of the first-order account, while residual semantic structure marks its boundary.
\end{itemize}

\section{Related Work}
\label{sec:related-work}

Prior work establishes that a modality gap exists, that modifying it can help or harm, and that commonly grouped interventions are not equivalent. What remains missing is a common geometric account of how the intervention and downstream readout jointly determine the observed outcome. We provide such an account.

\subsection{Modality-Gap Geometry}
\label{sec:rw-modality-gaps}

CLIP and SigLIP align paired samples without requiring the image and text marginals to coincide \citep{radford2021learning,zhai2023sigmoid}. Persistent separation has been connected to initialization, contrastive optimization, uniformity, information imbalance, temperature, and negative sampling \citep{liang2022mind,fahim2024contrastive,schrodi2025two,yaras2025explaining}. These studies measure different objects: centroid displacement is a first moment, whereas separability also reflects covariance, anisotropy, and modality-specific neighborhoods. Few coordinates dominate separation while structure persists \citep{schrodi2025two,levi2025double}; intra-modal interpretations remain contested \citep{mistretta2025cross,herzog2026reevaluating}. We study the dominant mean-separation component inside a frozen representation and distinguish it from the complete distribution.

\subsection{Representation Interventions}
\label{sec:rw-geometry-modification}

Methods described as gap removal implement different transformations. Offset subtraction translates embeddings; projection deletes a coordinate; projection plus normalization also rescales residuals \citep{liang2022mind,chowers2026bug}. Full modality-wise mean-centering, including the strong GR-CLIP baseline, removes complete centroids \citep{li2025closing}; dominant-component removal also appears in anisotropy correction \citep{mu2018allbutthetop}. Training-based alignment changes broader structure \citep{eslami2025mitigate,yamaguchi2025postpretraining,jiang2023latent}, while score calibration changes logits rather than embeddings \citep{zhu2023generalized}. Evidence for one intervention therefore does not transfer automatically to another.

\subsection{Downstream Consequences}
\label{sec:rw-downstream-consequences}

Classification, opposite-modality retrieval, and heterogeneous-index retrieval expose different candidate-varying quantities. Stronger alignment may help particular cross-modal retrieval settings while removing modality-specific information can harm classification or neighborhood quality \citep{liang2022mind,jiang2023latent,yaras2025explaining,yamaguchi2025postpretraining}. Mixed candidate pools additionally permit modality clusters to bias ranking, motivating mean-centering corrections \citep{li2025closing}. Candidate composition and normalization can thus change the sign of an intervention without contradiction.

\section{Readout-Dependent Modality-Gap Geometry}
\label{sec:gap-geometry}

We first define the dominant direction and distinguish four commonly conflated interventions. We then decompose the similarity score into modality-level, residual, and orthogonal components. The downstream role of each component depends on which quantities a task holds fixed and which it varies.

\subsection{Problem Setting and Direction Definitions}
\label{sec:dominant-gap-direction}

Let $x,t\in\mathbb{R}^d$ denote image and text embeddings in a shared representation space. Embeddings are L2-normalized whenever the score is cosine similarity. With modality centroids $\mu_{\mathrm{img}}$ and $\mu_{\mathrm{txt}}$, define
\begin{equation}
  g=\mu_{\mathrm{img}}-\mu_{\mathrm{txt}},\qquad
  \widehat g=\frac{g}{\lVert g\rVert_2}.
  \label{eq:dominant-gap-direction}
\end{equation}
The vector $\widehat g$ identifies the centroid-difference direction. We separately denote by $v_1$ the estimate-split leading principal direction of the joint image--text cloud. The experiments instantiate the intervention axis with $v_1$; its alignment with the independently estimated $\widehat g$ tests whether this stable leading direction organizes the image--text mean separation.

The score analysis holds for any unit direction $v$. For an embedding $z$,
\begin{equation}
  \beta_z=z^\top v,\qquad z_\perp=z-\beta_zv,\qquad
  z=z_\perp+\beta_zv.
  \label{eq:parallel-residual}
\end{equation}
Thus $z_\perp$ retains the geometry orthogonal to $v$. This characterization concerns concentration of the mean-separation component along $v_1$, not the complete semantic or distributional geometry.

\subsection{Four Non-Equivalent Interventions}
\label{sec:gap-operations}

The phrase \emph{gap removal} conflates transformations that preserve different information. Table~\ref{tab:operations} gives the conceptual map before the formal definitions.

\begin{center}
    \centering
    \small
    \setlength{\tabcolsep}{2pt}
    \begin{tabularx}{\columnwidth}{@{}lcccc>{\raggedright\arraybackslash\hyphenpenalty=10000}X@{}}
        \toprule
        Op. & \shortstack{Mean\\on $v$} & \shortstack{Sample\\coord.} & \shortstack{Pre-L2\\norm} & \shortstack{L2\\again} & \shortstack{Main\\consequence} \\
        \midrule
        O1 & Yes & No  & Yes & No  & Fixed modality translation \\
        O2 & Yes & Yes & Yes & No  & Deletes axis contribution \\
        O3 & Yes & Yes & Yes & Yes & Candidate-specific denominator \\
        O4 & Yes & No  & Yes & Yes & Full-centroid and cosine change \\
        \bottomrule
    \end{tabularx}
    \captionof{table}{\textbf{Non-equivalent embedding operations.} ``Sample coord.'' records deletion of the complete sample-specific coordinate on $v$; O4 instead removes the full modality centroid. Application side is specified separately.}
    \label{tab:operations}
\end{center}

Let $m\in\{\mathrm{img},\mathrm{txt}\}$ and $\bar\beta_m=\mathbb{E}_{z\sim m}[z^\top v]$. \textbf{O1}, fixed single-direction offset subtraction, is
\begin{equation}
  z'_m=z_m-\bar\beta_m v.
  \label{eq:fixed-rank-one-offset}
\end{equation}
It removes the modality mean along $v$ but retains sample residuals. \textbf{O2}, raw orthogonal projection, removes the complete sample coordinate,
\begin{equation}
  z'=z-(z^\top v)v=z_\perp.
  \label{eq:raw-gap-projection}
\end{equation}
It changes the norm without restoring unit length. \textbf{O3} projects and then L2-normalizes,
\begin{equation}
  z'=\frac{z_\perp}{\lVert z_\perp\rVert_2},\qquad
  \lVert z_\perp\rVert_2=\sqrt{1-(z^\top v)^2}
  \label{eq:project-and-normalize}
\end{equation}
for unit $z$, introducing a sample-dependent denominator. \textbf{O4}, full modality-wise mean-centering, is
\begin{equation}
  z'_m=z_m-\mathbb{E}[z_m],\qquad
  \widetilde z'_m=\frac{z'_m}{\lVert z'_m\rVert_2}
  \label{eq:full-modality-centering}
\end{equation}
when cosine is evaluated. O4 removes the full centroid and can modify directions orthogonal to $v$; it is not rank-one in general.

Score calibration is distinct from embedding modification because it changes the downstream readout without changing stored embeddings. The task effect therefore depends on the operation, application side, and any subsequent normalization.

\subsection{A Common Similarity-Score Decomposition}
\label{sec:score-decomposition}

For any unit $v$, decompose $x=x_\perp+\beta_xv$ and $t=t_\perp+\beta_tv$. Their dot-product score satisfies
\begin{equation}
  S(x,t)=x^\top t=x_\perp^\top t_\perp+\beta_x\beta_t.
  \label{eq:axis-score-decomposition}
\end{equation}
Write $\beta_x=\bar\beta_{\mathrm{img}}+r_x$ and $\beta_t=\bar\beta_{\mathrm{txt}}+r_t$. Then
\begin{align}
  S(x,t)&=S_\perp+A+B+C+D, \label{eq:abcd-decomposition}\\
  S_\perp&=x_\perp^\top t_\perp,\nonumber\\
  A&=\bar\beta_{\mathrm{img}}\bar\beta_{\mathrm{txt}},
  &B&=\bar\beta_{\mathrm{img}}r_t,\nonumber\\
  C&=\bar\beta_{\mathrm{txt}}r_x,
  &D&=r_xr_t.\nonumber
\end{align}
$A$ is a modality-level constant; $B$ and $C$ are first-order text- and image-residual terms; $D$ couples the sample coordinates; and $S_\perp$ retains similarity outside $v$. These algebraic labels are modality-based. In image-to-text retrieval, $B$ is candidate-varying and $C$ is query-side; in text-to-image retrieval, their functional roles exchange. Aggregated retrieval results therefore retain the algebraic $B/C$ labels but interpret candidate and query roles by direction.

O3 adds a distinct source of score change:
\begin{align}
  N_x&=\sqrt{1-\beta_x^2},&
  N_t&=\sqrt{1-\beta_t^2},\nonumber\\
  S_{\mathrm{proj\text{-}cos}}(x,t)
  &=\frac{x_\perp^\top t_\perp}{N_xN_t}.
  \label{eq:projected-cosine-score}
\end{align}
The numerator decomposition and candidate-dependent normalization are separate. The decomposition is exact, but functional sufficiency is empirical: near-tied scores can retain nearly identical numerical values while exchanging strict order.

\subsection{How Different Tasks Read the Same Geometry}
\label{sec:task-readouts}

Classification varies class prototypes, standard retrieval varies opposite-modality candidates, and mixed retrieval varies both candidate identity and modality, producing additive bias, scale distortion, and modality sorting, respectively.

\paragraph{Classification: additive class-dependent bias.}
Classification fixes an image query and varies text prototypes~$t_c$. Under the implemented query-only O1 intervention with $v=v_1$,
\begin{equation}
  (x-\bar\beta_{\mathrm{img}}v_1)^\top t_c
  =x^\top t_c-\bar\beta_{\mathrm{img}}\beta_{t_c}.
  \label{eq:classification-additive-identity}
\end{equation}
The second term is an additive class-dependent bias. This exact identity is specific to the fixed-offset construction and does not give O2--O4 the same interpretation.

\paragraph{Standard retrieval: candidate-specific scale distortion.}
A fixed query ranks opposite-modality candidates. After O3, each candidate has a different residual norm, producing a multiplicative calibration error. The implemented norm-power readout is
\begin{equation}
  S_\gamma(q,c)=
  \frac{q_\perp^\top c_\perp}{N_qN_c^{\,1-\gamma}}.
  \label{eq:norm-power-readout}
\end{equation}
For image-to-text retrieval, the geometry-derived first-order exponent is
\begin{equation}
  \gamma_t^{(1)} = 1 +
  \frac{|\bar\beta_{\mathrm{img}}|}
       {|\bar\beta_{\mathrm{txt}}|\,\bar S_{\perp,\mathrm{cos}}^{+}},
  \label{eq:gamma-first-order}
\end{equation}
where $\bar S_{\perp,\mathrm{cos}}^{+}$ is the estimate-split mean cosine between matched image--text residuals after projection and L2 normalization; the superscript $+$ denotes positive pairs. The expression is a simplified first-order approximation, with image and text roles exchanged for text-to-image. It predicts config-specific calibration rather than an exact optimum.

\paragraph{Mixed retrieval: modality-dependent candidate sorting.}
Image and text candidates share one index, so candidate modality varies within a ranking. Terms constant in a standard opposite-modality pool can then differ systematically across candidates, allowing the $v_1$ coordinate to sort by modality before semantic relevance.

One geometric account thus links four non-equivalent interventions and three readout mechanisms, with no task-independent sign. The modality gap is neither inherently beneficial nor harmful; its effect depends jointly on the operation and downstream readout.

\section{Experimental Setup}
\label{sec:experimental-protocol}

\subsection{Models, Tasks, and Evaluation Protocol}
\label{sec:models-datasets-tasks}
\label{sec:estimation-evaluation}

Every claim uses a frozen-geometry protocol: directions, coefficients, and thresholds are estimated on one split and applied unchanged to a disjoint evaluation split across five paired seeds. No evaluation outcome is used to choose a direction, coefficient, or threshold.

We use frozen final-layer embeddings from OpenAI CLIP ViT-B/32, ViT-B/16, and ViT-L/14 and Google SigLIP Base-P/16-224 and Large-P/16-256. Classification uses the two CLIP base encoders and SigLIP Base on CIFAR-10, CIFAR-100, ImageNet-1K, EuroSAT, and SVHN. Standard cross-modal retrieval uses Flickr30k with all five encoders and COCO-5K with the three base encoders, in both image-to-text and text-to-image directions. Binary mixed retrieval places image and text candidates from Flickr30k or COCO-5K in one shared index at image:text ratios of 25:75, 50:50, and 75:25. The matched O1--O4 cross-task comparison uses 16 mixed-retrieval configurations; the primary mechanism analysis uses 48 binary shared-index cells spanning eight encoder--dataset pairs, both query modalities, and three pool ratios. MixBench25/MSCOCO is evaluated separately because its documents may contain image, text, or both.

The held-out check uses OpenCLIP ConvNeXt-Base-W, whose architecture and training configuration are absent from development. Its geometry-only predictions are frozen before downstream observations are loaded.

Each task uses deterministic 50:50 estimate/evaluation partitions. Classification splits are stratified; retrieval partitions image identities and retains all associated captions; MixBench separates connected components of the relevance graph. The estimate split supplies centroids, coordinate means, residual-norm statistics, $v_1$, and fitted calibration references. The independently estimated centroid direction $\widehat g$ only measures alignment with $v_1$. Input embeddings are L2-normalized. Only operations that explicitly include post-intervention normalization normalize the embeddings again after transformation. Primary metrics are Top-1 accuracy, Mean Recall over Recall@1/5/10, and NDCG@10 for classification, standard retrieval, and mixed retrieval. Seed means and paired effects use 95\% Student-$t$ intervals; exponent correlations use a configuration bootstrap. The five-seed intervals quantify variability across the paired estimate/evaluation splits within the evaluated cells; they are not population intervals over encoder or dataset families.

\subsection{Interventions and Core Controls}
\label{sec:metrics-statistics}
\label{sec:baselines-controls}

We compare unchanged embeddings with O1 fixed offset subtraction, O2 raw projection, O3 projection followed by L2 normalization, and O4 full modality-wise mean-centering. The intervention direction for O1--O3 is $v_1$. The primary classification identity applies O1 to the image query; standard retrieval applies projection and normalization to both modalities; binary mixed retrieval applies each embedding operation to all indexed modalities. Matched PC2 and seeded random directions test whether an effect is specific to the dominant centroid-separation direction.

For standard retrieval, the geometry-derived first-order exponent and estimate-split fitted exponent are frozen before evaluation. A shared constant tests whether aggregate recovery alone is sufficient. These methods recalibrate scores and do not alter stored embeddings. The held-out protocol also freezes geometry-only predictions before downstream observations, separating qualitative transfer from post-hoc quantitative fit.

\section{From Gap Geometry to Task-Dependent Roles}
\label{sec:results}

\begin{figure*}[t]
  \centering
  \includegraphics[width=0.98\textwidth]{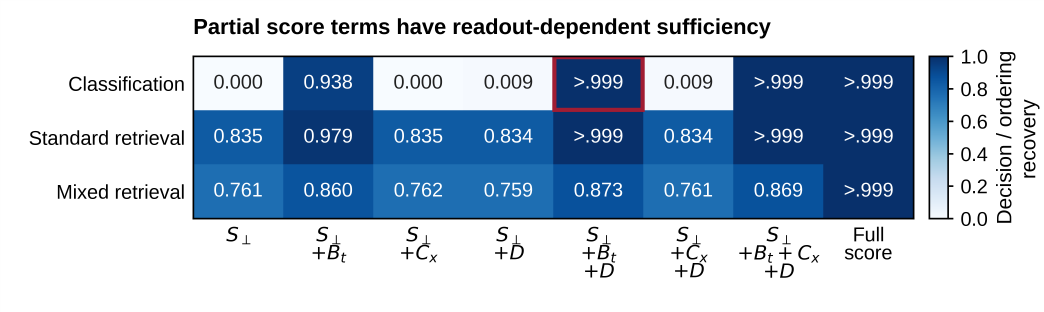}
  \caption{\textbf{Score-term sufficiency is readout-dependent.} Cells show five-seed mean prediction or ordering recovery, rather than numerical score error. The algebraic $B/C$ labels are image/text-based: $B$ is candidate-varying for image-to-text retrieval and $C$ for text-to-image, so the retrieval row aggregates direction-specific functional roles. The full score gives exact numerical reconstruction within tolerance; near ties can still change strict decisions.}
  \label{fig:score-reconstruction}
\end{figure*}

\begin{figure*}[t]
  \centering
  \includegraphics[width=0.98\textwidth]{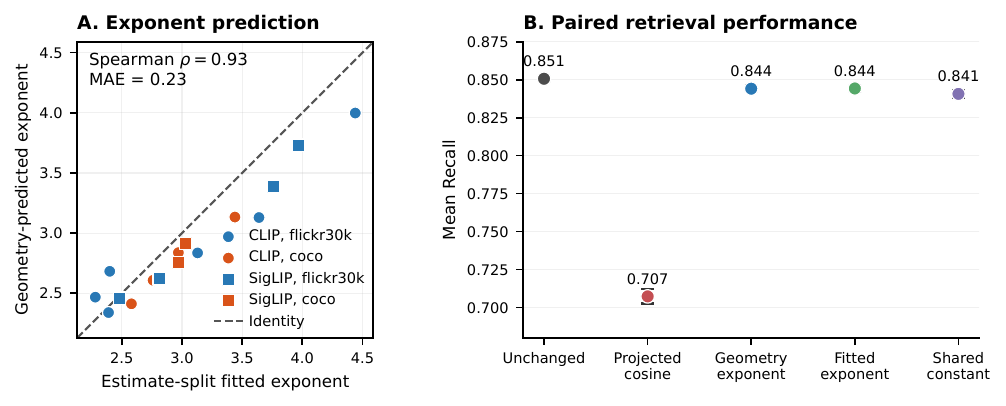}
  \caption{\textbf{Geometry predicts configuration-specific scale calibration.} \textbf{A:} predicted versus estimate-split fitted exponents for 16 configurations; Spearman $\rho=0.930$ with a 2,000-resample bootstrap interval. \textbf{B:} evaluation Mean Recall over five paired seeds; error bars show 95\% Student-$t$ intervals. The fitted exponent is diagnostic and the shared constant is configuration-invariant.}
  \label{fig:gamma-prediction}
\end{figure*}

\subsection{Mean Separation Concentrates Along One Stable Direction}
\label{sec:rank-one-results}

The image--text centroid separation is, to a close approximation, concentrated along one stable direction. Across 23 encoder--dataset cells, the estimate-split $v_1$ captured 94.4--99.9\% of the squared image--text mean separation on the disjoint evaluation split. Its estimate/evaluation cosine ranged from 0.9980 to above 0.9999, and $|\widehat g^\top v_1|$ ranged from 0.9718 to above 0.9999. Figure~\ref{fig:empirical-motivation} includes the weakest cell and representative CLIP and SigLIP cells. Modality AUC was 1.0 along $v_1$, compared with 0.437--0.781 for PC2 and 0.500 on average over 460 seeded random axes. Cross-dataset stability held in 62 of 63 comparisons, with one lower-bound miss. This approximately rank-one characterization concerns the mean-separation component, not the complete multimodal distribution.

\subsection{Non-Equivalent Interventions Produce Different Outcomes}
\label{sec:operation-results}

Panel I of Figure~\ref{fig:empirical-motivation} compares O1--O4 on matched development grids. O1 and O2 reduced classification Top-1 by 2.06 and 2.13 points and standard-retrieval Mean Recall by 0.069 and 0.071, while increasing mixed NDCG@10 by 0.132 and 0.135. O3 applies the same coordinate deletion as O2 and then normalizes each residual; its losses were larger in classification (9.14 points) and standard retrieval (0.143), while mixed NDCG still increased by 0.122. O4 improved classification by 2.47 points and mixed NDCG by 0.169, with a standard-retrieval change of $-0.014$. These mixed values use the matched 16-configuration operation grid and are distinct from the 48-cell binary analysis in Section~\ref{sec:mixed-results}.

O1 removes a modality-level offset while retaining sample variation along $v_1$; O2 removes the full sample coordinate; O3 adds sample-dependent normalization; and O4 removes the full centroid. O1 and O2 nevertheless leave nearly identical mean residual gaps (0.05821 and 0.05817). Similar gap reduction therefore does not imply equivalent score or task behavior.

These geometric modifications can improve one readout while degrading a different one. Remaining centroid distance alone cannot determine whether gap modification is beneficial.

\subsection{Readout-Dependent Score Reconstruction}
\label{sec:reconstruction-results}

Equation~\eqref{eq:abcd-decomposition} reconstructed every tested score within numerical tolerance, but partial terms preserved decisions differently across readouts (Figure~\ref{fig:score-reconstruction}).

For classification, $S_\perp+B$ recovered 0.938 of prediction changes and $S_\perp+B+D$ exceeded 0.9999. Standard-retrieval ordering recovery rose from 0.835 with $S_\perp$ to 0.979 after adding $B$, while the projected-cosine denominator remained a separate source of change. Mixed retrieval was least compressible: $S_\perp+B+D$ recovered 0.873. High average recovery does not guarantee all-seed agreement because near-tied predictions or rankings can change under small numerical differences. The decomposition is exact, but functional sufficiency remains task-dependent.

\subsection{Classification: Additive Class-Dependent Bias}
\label{sec:classification-results}

Under query-only O1, Eq.~\eqref{eq:classification-additive-identity} gives the complete class-dependent score shift exactly. The recovery percentages below answer a different question: whether selected decomposition terms reproduce strict predictions. The fixed offset changed 27.07\% of predictions and reduced cell-averaged Top-1 from 60.71\% to 58.65\%. This net 2.06-point reduction conceals substantial bidirectional churn: 7.82\% of examples change from correct to incorrect, while 5.76\% change from incorrect to correct. The additive component is therefore not uniformly harmful; its effect depends on alignment with the class prototypes. Term $B$ recovered 93.84\% of prediction changes (95\% CI 93.68--94.00\%); adding $D$ raised recovery above 99.99\%, with maximum score error $7.4\times10^{-7}$. A matched PC2 reinjection recovered 0.626 of the score response versus 1.000 for the measured direction. Under the fixed-query O1 construction, the dominant coordinate acts as an additive class-dependent bias; the identity does not extend unchanged to O2--O4.

\subsection{Standard Retrieval: Candidate-Specific Scale Distortion}
\label{sec:retrieval-results}

Projection followed by normalization reduced Mean Recall from 0.8506 to 0.7073 across 16 encoder--dataset--direction configurations. The geometry-derived exponent raised it to 0.8441, recovering 95.4\% of the lost performance (Figure~\ref{fig:gamma-prediction}).

The predicted exponent tracked the fitted value with Spearman $\rho=0.930$ (95\% CI 0.730--0.991) and MAE 0.230. A shared constant reached 0.8407 Mean Recall, but its exponent MAE was 0.486; geometry was closer to the fitted value in 12 of 16 configurations. A shared constant therefore recovers much of the average loss, but cannot predict which configurations require stronger or weaker correction. Transfer remained uneven, and the full first-order predictor had higher error (MAE 0.534). The exponent is therefore not a universal cross-dataset calibrator, but a configuration-sensitive first-order explanation of why projection followed by normalization distorts candidate rankings.

\subsection{Mixed Retrieval: Modality-Dependent Candidate Sorting}
\label{sec:mixed-results}

The primary mixed result uses 48 binary shared-index cells: eight encoder--dataset pairs, both query modalities, and three image:text pool ratios. Each cell uses five paired seeds.

Removing $v_1$ transformed shared-index retrieval: NDCG@10 rose from 0.0002 to 0.2160, while pool-adjusted modality skew fell from 0.9999 to 0.7952. Relevance is assigned to paired opposite-modality items while image and text candidates share the pool; unchanged scores rank same-modality candidates ahead of the relevant cross-modal items, producing the near-zero baseline. Removing PC2 or seeded random directions left performance near the unchanged system, whereas O4 reached 0.2417. The dominant direction therefore acts as a strong modality-sorting component within the shared index.

Table~\ref{tab:mixed-retrieval-results} compares the dominant-direction intervention with matched non-gap controls and full centering.
\begin{table}[t]
  \centering
  \footnotesize
  \setlength{\tabcolsep}{1.2pt}
  \begin{tabularx}{\columnwidth}{@{}>{\raggedright\arraybackslash\hyphenpenalty=10000}Xcccc@{}}
    \toprule
    Method & \shortstack{Binary\\NDCG@10 $\uparrow$} & \shortstack{Pool-adj.\\skew $\downarrow$} & \shortstack{Within-mod.\\$\rho\uparrow$} & \shortstack{Standard ret.\\$\Delta$MR} \\
    \midrule
    Unchanged & 0.0002 & 0.9999 & 1.000 & 0.000 \\
    O2: remove $v_1$ & 0.2160 & \textbf{0.7952} & 0.896 & $-0.071$ \\
    Remove PC2 & 0.0013 & 0.9993 & 0.872 & --- \\
    Remove seeded random & 0.0002 & 0.9999 & 1.000 & --- \\
    O4: full centering & \textbf{0.2417} & 0.8175 & 0.796 & $-0.014$ \\
    \bottomrule
  \end{tabularx}
  \caption{\textbf{Binary shared-index retrieval is direction-specific.} Means over 48 cells across eight Flickr30k/COCO-5K encoder--dataset pairs, both query modalities, and image:text ratios 25:75, 50:50, and 75:25; five paired seeds per cell; MixBench is excluded. Near-zero unchanged NDCG accompanies severe modality-dominated ranking. Standard-retrieval $\Delta$MR is from a separate matched 16-configuration evaluation.}
  \label{tab:mixed-retrieval-results}
\end{table}

O2 preserved more within-modality order ($\rho=0.896$) than O4 ($\rho=0.796$), but cost 0.071 standard-retrieval Mean Recall versus 0.014 for O4 on a separate matched standard-retrieval grid. This contrast clarifies that the binary shared-index gains do not come for free: rank-one removal better preserves within-modality structure, but it also produces a larger trade-off on the conventional opposite-modality retrieval task. Thus, the same intervention can substantially improve cross-modal competition in a shared index while harming standard retrieval, and the preferred operation depends on which readout is ultimately being used.

MixBench is a separate text-query construction with image, text, and multimodal documents, so it complements rather than duplicates the binary shared-index analysis. Its NDCG@10 was 0.4543 unchanged, 0.6612 after $v_1$ removal, and 0.6892 after full centering; content-type filtering control reached only 0.2996. The primary ranking criterion improved under both interventions, but retrieval-score modality AUC moved farther from chance after centering. Together with the binary controls, these results support the interpretation that the dominant direction acts as a modality-dependent sorting component, while stopping short of claiming complete restoration of mixed-modal retrieval.

\section{Scope and Boundary of the Rank-One Account}
\label{sec:mechanism-validation}

A mechanistic account is useful only when its domain of validity is explicit. We examine three boundaries: direction specificity, architectural transfer, and residual structure.

\begin{table}[htbp]
  \centering
  \begin{tabular}{@{}p{0.68\columnwidth}r@{}}
    \toprule
    Check & Evidence \\
    \midrule
    \multicolumn{2}{@{}l}{\textit{A. Held-out checkpoint}} \\
    Scalar effect directions & 7/7 \\
    Operation-effect signs & 10/12 \\
    Exact task ordering & 1/3 \\
    Frozen interval coverage & 1/7 \\
    Scalar / operation-effect MAE & 0.257 / 0.092 \\
    \midrule
    \multicolumn{2}{@{}l}{\textit{B. Residual boundary}} \\
    Centroid removed; distribution persists & 21/23 \\
    Residual subspace multidimensional & 23/23 \\
    Stable non-modality retrieval direction & 8/8 \\
    Classification median / random p95 & 0.0707 / 0.0015 \\
    Retrieval median / random p95 & 0.0083 / 0.0012 \\
    \bottomrule
  \end{tabular}
  \caption{\textbf{Transfer and boundary. } Frozen held-out prediction (A) and residual structure after centering (B).}
  \label{tab:transfer-boundary}
\end{table}

\FloatBarrier
\subsection{Direction Specificity}
\label{sec:reinjection}

Exact unit-strength reinjection of $v_1$ with its original coordinates is an expected reconstruction check. The informative tests change the direction, coefficients, angle, or strength while preserving the intervention budget. Across matched wrong-direction and coefficient controls, 96.3\% satisfied the predefined specificity criterion, although generic low-dimensional reinjection also recovered part of the score variation. Thus, direction and coefficient identity both matter.

\subsection{Held-Out OpenCLIP ConvNeXt Results}
\label{sec:heldout-prediction}

For a held-out OpenCLIP ConvNeXt-Base-W checkpoint, geometry-only predictions were frozen before downstream evaluation. Qualitative transfer was strong: all seven scalar effect directions and 10 of 12 operation-effect signs were predicted correctly on an architecture and training configuration absent from development. Quantitative transfer was weaker: only one frozen interval covered its observation, and complete operation ordering transferred in one of three tasks (Table~\ref{tab:transfer-boundary}). The first-moment account therefore captures qualitative mechanism more reliably than architecture-specific magnitude.

\subsection{Residual Structure Beyond the Rank-One Account}
\label{sec:boundary}

After full mean-centering, distributional modality separation persisted in 21 of 23 tested cells, while residual PCA was multidimensional in all 23. Orthogonal semantic directions had median absolute effects of 0.0707 versus a random-direction 95th percentile of 0.0015 in classification and 0.0083 versus 0.0012 in retrieval; stable retrieval directions exceeded random controls in all eight cells. These residual directions remain task-relevant while exhibiting near-chance modality discrimination. They therefore represent semantic structure orthogonal to the dominant modality-separation component, rather than the original modality gap reappearing. This distinction separates removal of the dominant first-moment offset from collapse of the broader representation geometry. Mean-centering can eliminate the principal modality-separating component while leaving structured variation that still affects downstream decisions. The dominant mean-separation component is important but does not exhaust multimodal geometry.

\section{Discussion and Limitations}
\label{sec:discussion}

The results reconcile conflicting reports because interventions described as gap modification preserve different information, while candidate composition determines which changed terms affect a downstream readout. Future studies should report: (1) the intervention direction, (2) application side, (3) post-intervention normalization, (4) candidate-modality composition, and (5) downstream readout. Full centering is a strong practical baseline but also modifies orthogonal directions; score calibration changes the readout rather than stored embeddings.
Conceptually, the modality gap is neither uniformly a defect nor uniformly useful. Its functional role emerges from the interaction between representation geometry and the readout that consumes it. Consequently, no task-independent operation is universally preferable: projection isolates direction-specific effects, projection plus normalization exposes candidate-scale distortion, and full centering remains a strong practical intervention when mixed-modality competition is central.

Our scope is frozen CLIP and SigLIP dual encoders, final-layer embeddings, and similarity-based tasks. The retrieval approximation transfers unevenly, held-out signs transfer better than magnitudes, and residual multidimensional structure remains after centering. The account does not extend directly to fused or generative models and does not replace higher-order analysis. Within this scope, the dominant mean-separation component provides a coherent first-order account of why gap modification helps or harms different readouts.

\section{Conclusion}
\label{sec:conclusion}

The modality gap has no single task-independent consequence or remedy. Its dominant mean-separation component is approximately rank-one, but its downstream effect is determined by the interaction between the intervention and the readout. Practitioners should therefore diagnose the readout before choosing an intervention and report both when evaluating gap modification.
Our results suggest that geometry-aware intervention design should guide future development of vision-language models.

\bibliographystyle{plainnat}
\bibliography{references}

\end{document}